\documentclass[10pt,twocolumn,letterpaper]{article}
\usepackage[accsupp]{axessibility}
\usepackage{iccv}              

\usepackage[pagebackref,breaklinks,colorlinks,allcolors=iccvblue]{hyperref}
\usepackage{arydshln}
\usepackage{multirow}
\usepackage{makecell} 
\usepackage[table]{xcolor}
\usepackage{graphicx} 
\usepackage{amsmath} 
\usepackage{color}
\usepackage{adjustbox}
\usepackage{amssymb}
\usepackage{array}
\usepackage{pifont}
\usepackage{tabularx}
\def\paperID{10096} 
\def\confName{ICCV}
\def\confYear{2025}

\def \Ourbench {STI-Bench}

\definecolor{iccvblue}{rgb}{0.21,0.49,0.74}
\definecolor{lightblue}{RGB}{173,216,230}
\definecolor{navyblue}{HTML}{0071BC}

\definecolor{orange}{HTML}{FFA64D} 
\definecolor{light-orange}{HTML}{FFD4A3} 
\definecolor{my_green}{RGB}{51,102,0}
\definecolor{my_red}{RGB}{204, 0, 0}
\renewcommand{\checkmark}{\textcolor{my_green}{\ding{51}}} 
\newcommand{\crossmark}{\textcolor{my_red}{\ding{55}}} 

\title{{\Ourbench}:\\ Are MLLMs Ready for Precise Spatial-Temporal World Understanding?}

\author{
  Yun Li$^{1,2}$, 
  Yiming Zhang$^{1,3}$, 
  Tao Lin$^{1}$, 
  XiangRui Liu$^{1,4}$, 
  Wenxiao Cai$^{5}$, 
  Zheng Liu$^{4}$, 
  Bo Zhao$^{1}$\\
  $^1$School of AI, Shanghai Jiao Tong University\\
  $^2$China University of Geosciences,
  $^3$Nanyang Technological University,   
  $^4$BAAI,
  $^5$Stanford University\\
  Corresponding to \texttt{$<$bo.zhao@sjtu.edu.cn$>$}
}

\begin{document}
\maketitle

\begin{abstract}
The use of Multimodal Large Language Models (MLLMs) as an end-to-end solution for Embodied AI and Autonomous Driving has become a prevailing trend. While MLLMs have been extensively studied for visual semantic understanding tasks, their ability to perform precise and quantitative spatial-temporal understanding in real-world applications remains largely unexamined, leading to uncertain prospects. To evaluate models' Spatial-Temporal Intelligence, we introduce {\Ourbench}, a benchmark designed to evaluate MLLMs' spatial-temporal understanding through challenging tasks such as estimating and predicting the appearance, pose, displacement, and motion of objects. Our benchmark encompasses a wide range of robot and vehicle operations across outdoor, indoor, and desktop scenarios. 
The extensive experiments reveal that the state-of-the-art MLLMs still struggle in real-world spatial-temporal understanding, especially in tasks requiring precise distance estimation and motion analysis. Paper Page: \href{https://mint-sjtu.github.io/STI-Bench.io/}{https://mint-sjtu.github.io/STI-Bench.io/}

\end{abstract}

\begin{figure}[t]
    \centering
    \includegraphics[width=\columnwidth]{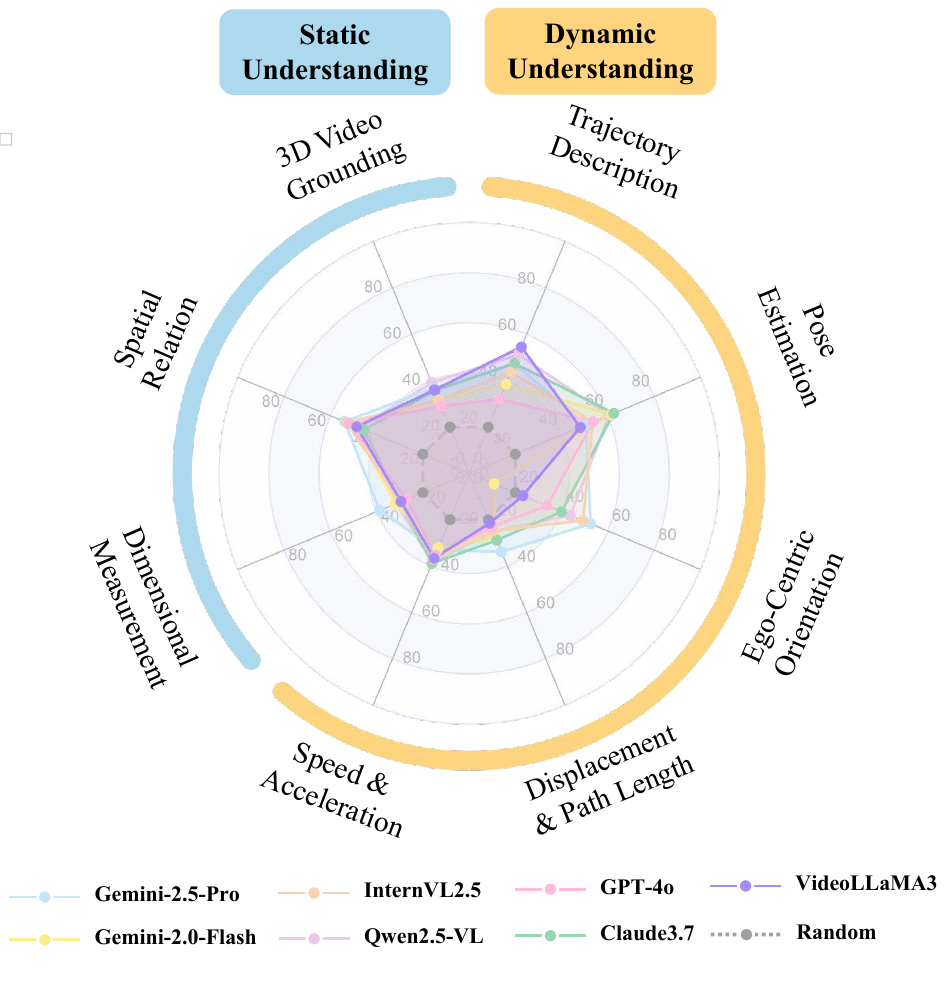}
    \caption{We evaluate state-of-the-art MLLMs on STI-Bench for precise and quantitative spatial-temporal understanding using video inputs. Results indicate the significant challenge in all tasks.}
    \label{fig:radar}
\end{figure}

\begin{figure*}[h!]
    \centering
    \includegraphics[width=\textwidth]{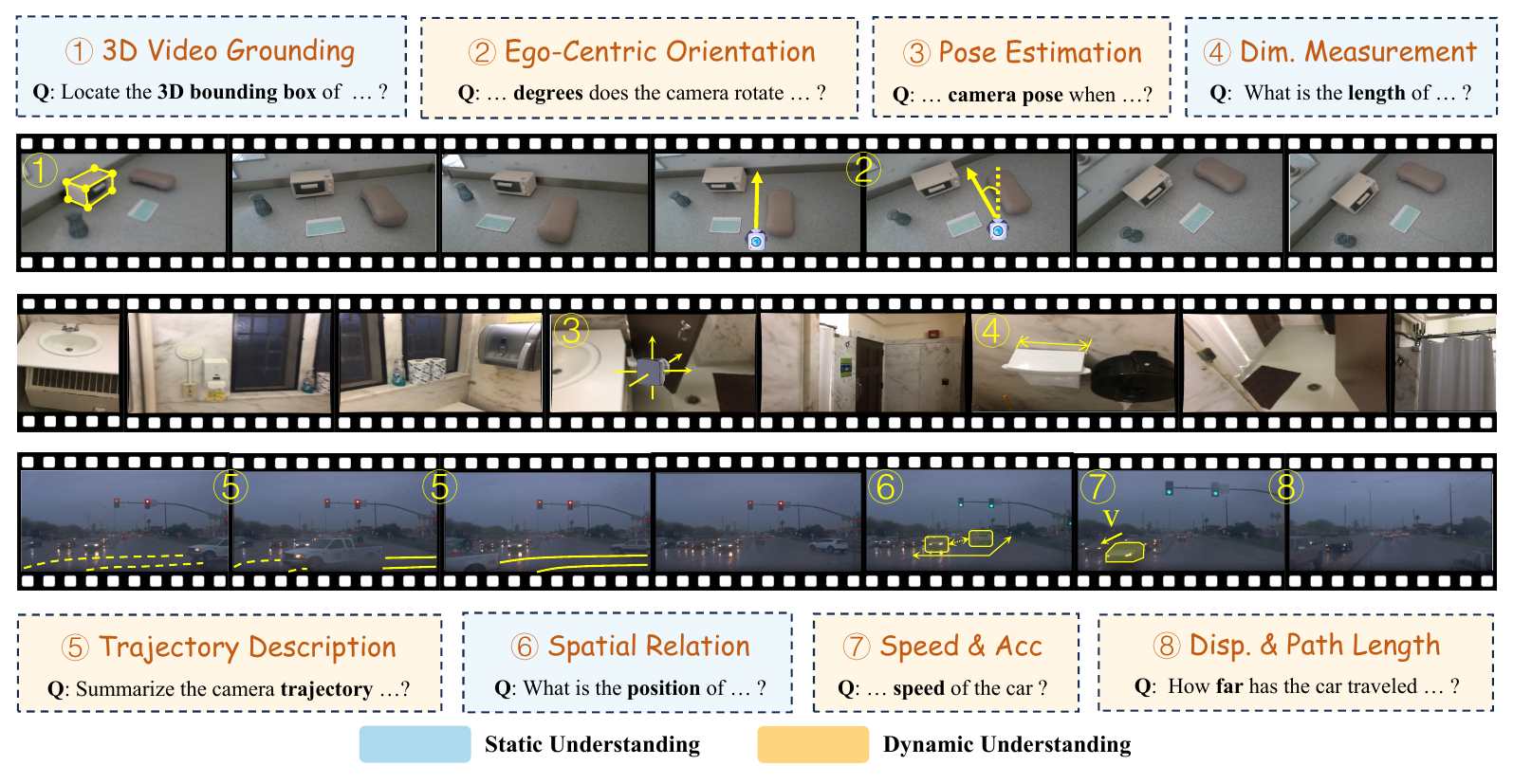}
    \caption{\textbf{Overview of STI-Bench.} We selected the most representative videos from each dataset scene and provided a few simple questions for demonstration.  }
    \label{fig:cover}
\end{figure*}

\section{Introduction}
Recent advances in Multimodal Large Language Models (MLLMs)~\cite{alayrac2022flamingo,chen2024internvl,bai2023qwen,team2023gemini,team2024gemini,touvron2023llama,lin2023video,shu2024video,zhang2025videollama,yang2024qwen2}
have positioned them as powerful tools for numerous vision and multimodal tasks.
MLLMs achieve impressive performance in general Visual Question Answering tasks~\cite{antol2015vqa}, which mainly focus on 2D visual perception and semantic question answering~\cite{fang2024mmbench,li2024mvbench,wang2024lvbench,wu2025longvideobench,fu2024video,zhou2024mlvu,hu2025video}.
Beyond 2D visual perception, MLLMs are increasingly employed as end-to-end solutions for Embodied AI~\cite{li2024behavior,mees2022calvin,li2024embodied,yang2025embodiedbench,cheng2025embodiedeval, cai2024spatialbot,brohan2023rt,black2024pi_0} and Autonomous Driving~\cite{xu2024drivegpt4, yuan2024rag,hwang2024emma,xu2024vlmad}. Such tasks require MLLMs to understand 3D space and time, and predict optimal manipulation strategies for robotic and vehicular systems. Despite these explorations, the question remains: Are MLLMs ready for precise spatial-temporal world understanding?

To answer this question, we propose a \textbf{S}patial-\textbf{T}emporal \textbf{I}ntelligence \textbf{Bench}mark (\textbf{STI-Bench}), designed to evaluate MLLMs' spatial-temporal world understanding capability. We evaluate MLLMs using single video or multiple images as input instead of 3D point clouds. The main reasons are: 1) the majority of state-of-the-art models, e.g., GPT-4o \cite{gpt4o} and Gemini \cite{team2024gemini}, accept images or video as input rather than 3D point clouds; 2) Videos are more frequently used in daily life and contain sufficient information to infer the spatial-temporal environment.

STI-Bench contains 300 videos and over 2,000 QA pairs,
covering three major scenarios: \emph{Desktop}, \emph{Indoor},
and \emph{Outdoor}.
The videos are sourced from Waymo~\cite{Sun_2020_CVPR}, ScanNet~\cite{dai2017scannet}, and Omni6DPose~\cite{zhang2024omni6dpose} respectively, encompassing a broad spectrum of real-world environments. As illustrated in Figure~\ref{fig:cover}, we design eight distinct tasks to evaluate models' ability in static spatial measurement and grounding, and dynamic tasks including speed, acceleration and trajectory estimation.

Through extensive experiments as illustrated in Figure~\ref{fig:radar}, we observe that even the most advanced MLLMs struggle with real-world spatial-temporal understanding, especially in tasks requiring precise distance estimation and motion analysis. Our error analysis reveals three fundamental limitations: inaccurate spatial quantification, flawed temporal dynamics understanding, and weak cross-modal grounding and integration.
These insights highlight the significant challenges MLLMs face in precisely understanding spatial-temporal information from videos. We believe STI-Bench will serve as an important touchstone that guides the community to develop better MLLMs for Embodied AI, Autonomous Driving tasks and beyond.

In summary, our main contributions include:
\begin{itemize}
\item We present STI-Bench, comprising over 300 videos and over 2,000 tailored questions across outdoor, indoor, and desktop scenarios, providing a systematic quantitative assessment of MLLMs' spatial-temporal understanding capabilities.
\item We conduct an in-depth study of state-of-the-art video-based MLLMs on STI-Bench, identify key error patterns in spatial-temporal reasoning, and provide empirical insights that can help the community develop more reliable MLLMs for embodied applications.
\end{itemize}

\begin{table*}[ht!]
\centering
\small

\renewcommand{\arraystretch}{1.0} 
\begin{adjustbox}{max width=\textwidth}
\setlength\tabcolsep{3pt}
\begin{tabular}{lccccccccccccccc}
\toprule
\multirow{2}{*}{\textbf {Benchmark}} &
\multirow{2}{*}{\textbf{QA Pairs}} & \multirow{2}{*}{\textbf{Data}} & \multirow{2}{*}{\textbf{Env.}} & \multicolumn{3}{c}{\textbf{Scene}} & \multicolumn{2}{c}{\textbf{View}} & \multicolumn{2}{c}{\textbf{Evaluation}} & \multicolumn{4}{c}{\textbf{Spatial-Temporal}}  \\
\cmidrule(lr){5-7} \cmidrule(lr){8-9} \cmidrule(lr){10-11} \cmidrule(lr){12-15}
&  &  &  & \textbf{D} & \textbf{I} & \textbf{O} & \textbf{Ego} & \textbf{Allo.} & \textbf{Num.} & \textbf{Desc.} & \textbf{Dist.} & \textbf{Dir.} & \textbf{Vel.} & \textbf{Traj.} \\
\midrule
SAT \cite{ray2024sat} & 218k & \textbf{I} & \textbf{S} & \crossmark & \checkmark & \crossmark & \checkmark & \checkmark & \checkmark & \checkmark & \crossmark & \crossmark & \crossmark & \crossmark \\
VSI-Bench \cite{yang2024thinking} & 5,156 & \textbf{V} & \textbf{R} & \checkmark & \crossmark & \crossmark & \checkmark & \checkmark & \checkmark & \checkmark & \checkmark & \crossmark & \crossmark & \checkmark \\
EmbSpatial-Bench \cite{du2024embspatial} & 3,640 & \textbf{I} & \textbf{R} & \crossmark  & \checkmark & \crossmark & \checkmark & \crossmark & \crossmark  & \checkmark & \crossmark & \crossmark & \crossmark & \crossmark \\ 
EmbodiedAgentInterface \cite{li2024embodied} & 448 & - & \textbf{S} & \crossmark  & \checkmark & \crossmark & \checkmark & \crossmark & - & -  & \crossmark & \crossmark & \crossmark & \crossmark \\

EmbodiedEval \cite{cheng2025embodiedeval} & 328 & \textbf{I/V} & \textbf{S} & \crossmark  & \checkmark & \checkmark & \checkmark & \crossmark & - & -  & \crossmark & \crossmark & \crossmark & \crossmark \\
EmbodiedBench \cite{yang2025embodiedbench} & 1,128 & \textbf{I} & \textbf{S} & \crossmark  & \checkmark & \checkmark & \checkmark & \crossmark & - & - & \crossmark & \crossmark & \crossmark & \crossmark \\

\midrule

WorldSense \cite{benchekroun2023worldsense} & 3,172 & \textbf{V} & \textbf{R} & \checkmark & \checkmark & \checkmark & \checkmark & \checkmark & \crossmark & \checkmark & \crossmark & \crossmark & \crossmark & \crossmark \\ 
MLVU \cite{zhou2024mlvu} & 3,102 & \textbf{V} & \textbf{R} & \checkmark & \checkmark & \checkmark  & \checkmark & \checkmark & \crossmark & \checkmark & \crossmark & \crossmark & \crossmark & \crossmark \\
Video-MMMU \cite{hu2025video}& 300 & \textbf{V} & \textbf{S} & \crossmark & \crossmark & \crossmark & \crossmark & \crossmark & \crossmark & \checkmark & \crossmark & \crossmark & \crossmark & \crossmark \\

\midrule
\Ourbench & 2,064 & \textbf{V} & \textbf{R} & \checkmark & \checkmark & \checkmark & \checkmark & \checkmark & \checkmark & \checkmark & \checkmark & \checkmark & \checkmark & \checkmark \\
\bottomrule
\end{tabular}
\end{adjustbox}
\vspace{-0.2cm}
\caption{\textbf{Comparison of {\Ourbench} with existing benchmarks.} \textbf{Data} represents the source of our QA data, where \textbf{V} stands for Video and \textbf{I} stands for Image. \textbf{Env.} indicates the environment in which the data is generated, where \textbf{S} represents Simulation and \textbf{R} represents Real. The two columns under \textbf{View} indicate whether the dataset includes Ego-centric and Allocentric perspectives. The two columns under \textbf{Evaluation} specify whether the ground truth is presented in numerical or textual form. The four columns under \textbf{Spatial-Temporal} indicate whether the benchmark evaluates spatial distance, direction (with angular precision), velocity, or a precise and comprehensive trajectory description.}
\vspace{-0.5cm}
\label{tab:comparison}
\end{table*}

\section{Related Work}
\subsection{Multimodal Large Language Models}
MLLMs have achieved groundbreaking performance in visual understanding~\cite{alayrac2022flamingo, bai2023qwen, chen2024internvl, team2024gemini}, leveraging large language models (LLMs)~\cite{yang2024qwen2,touvron2023llama, team2023gemini} and visual encoders. Recent advancements have extended multimodal learning to video understanding.
Classical works include VideoChat\cite{li2023videochat}, which enables interactive video-based dialogue by integrating multimodal understanding. Subsequent models like Video-LLaVA\cite{lin2024videollavalearningunitedvisual} enhance visual-language alignment through large-scale vision-language pretraining, extending LLaVA\cite{liu2023llava}'s capability to process video inputs effectively. Recent works like Qwen2.5-VL \cite{yang2024qwen2} excel in long-video understanding and temporal localization by incorporating absolute temporal encoding, enabling models to capture relationships among video frames more effectively.

\subsection{Spatial Understanding with MLLMs}
Video MLLMs have focused heavily on semantic understanding. However, spatial understanding remains a significant challenge, inspiring recent contributions \cite{cai2024spatialbot, chen2024spatialvlm, cheng2024spatialrgpt}. This progress represents a significant step toward developing world models and embodied agents.
Recent advancements in embodied intelligence have explored integrating large-scale MLLMs into robotic control, enabling better generalization and semantic reasoning. RT-2~\cite{brohan2023rt} introduces a vision-language-action framework that transfers web-scale knowledge to robotic control by representing actions as tokens alongside visual and language data. Building on this idea, GR-2~\cite{cheang2024gr} extends generalist robot control across diverse embodiments using a Transformer-based architecture. Further refining this approach, $\pi_{0}$~\cite{black2024pi_0} incorporates a flow-matching mechanism to generate continuous, precise action trajectories, enhancing fine-grained manipulation skills.

\subsection{Video Benchmarks for MLLM}
Recently, multiple benchmarks \cite{wu2025longvideobench, wang2024lvbench, zhou2024mlvu, fu2024video} have emerged for comprehensively evaluating MLLMs' ability in (long) video understanding, especially visual perception and semantic reasoning through Video Question Answering. LongVideoBench~\cite{wu2025longvideobench} and LVBench~\cite{wang2024lvbench} focus on long video understanding, while Video-MME~\cite{fu2024video} and MMBench-Video~\cite{fang2024mmbench} comprehensively evaluate MLLMs across various video-related tasks.
Existing benchmarks primarily focus on high-level semantic understanding, such as entity recognition and event understanding, largely confined to a temporal extension of 2D image understanding, lacking precise 3D spatial and temporal reasoning of physical quantities.
Recent works such as VSI-Bench~\cite{yang2024thinking} have introduced visual-spatial intelligence tasks for MLLMs, where models are required to provide numerical answers in certain scenarios.
However, as illustrated in Table~\ref{tab:comparison}, the limited inclusion of scenes and spatial-temporal tasks restricts their ability to capture the complexities of the real physical world.
In contrast, {\Ourbench} comprehensively evaluates models' ability in precise spatial-temporal understanding through tasks of static spatial measurement and physical motion understanding in Desktop, Indoor and Outdoor scenarios.


\begin{figure}[t]
    \centering
    \includegraphics[width=0.9\columnwidth]{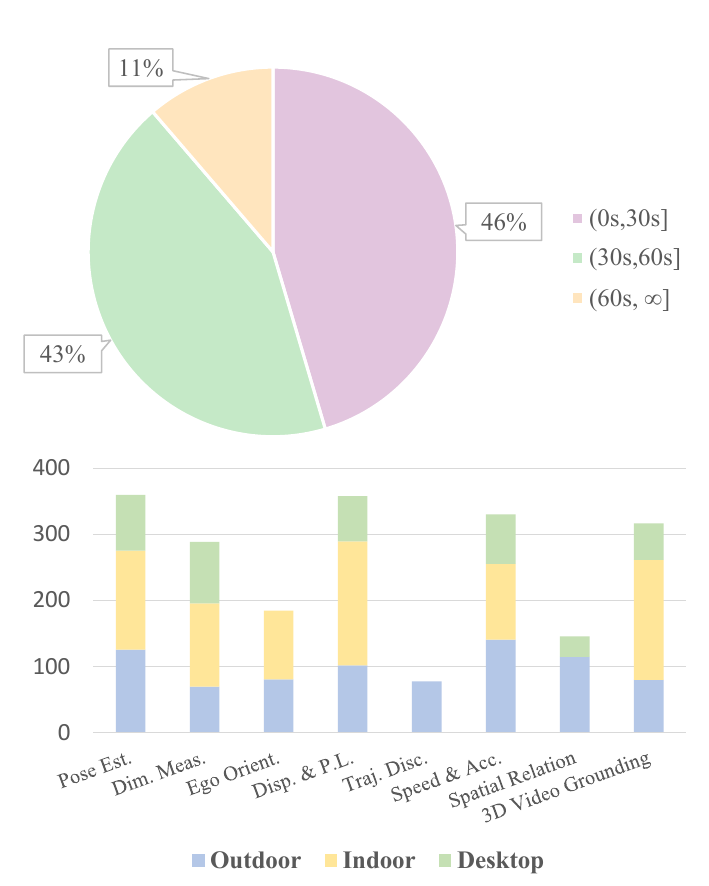}
    \caption{\textbf{Benchmark Statistics.} Top: Video length distribution across different categories and datasets. Bottom: The number of questions contributed by each dataset for evaluating different capabilities.}
    \label{fig:statistic}
\end{figure}

\section{STI-Bench}
\begin{figure*}[t]
    \centering
    \includegraphics[width=0.9\textwidth]{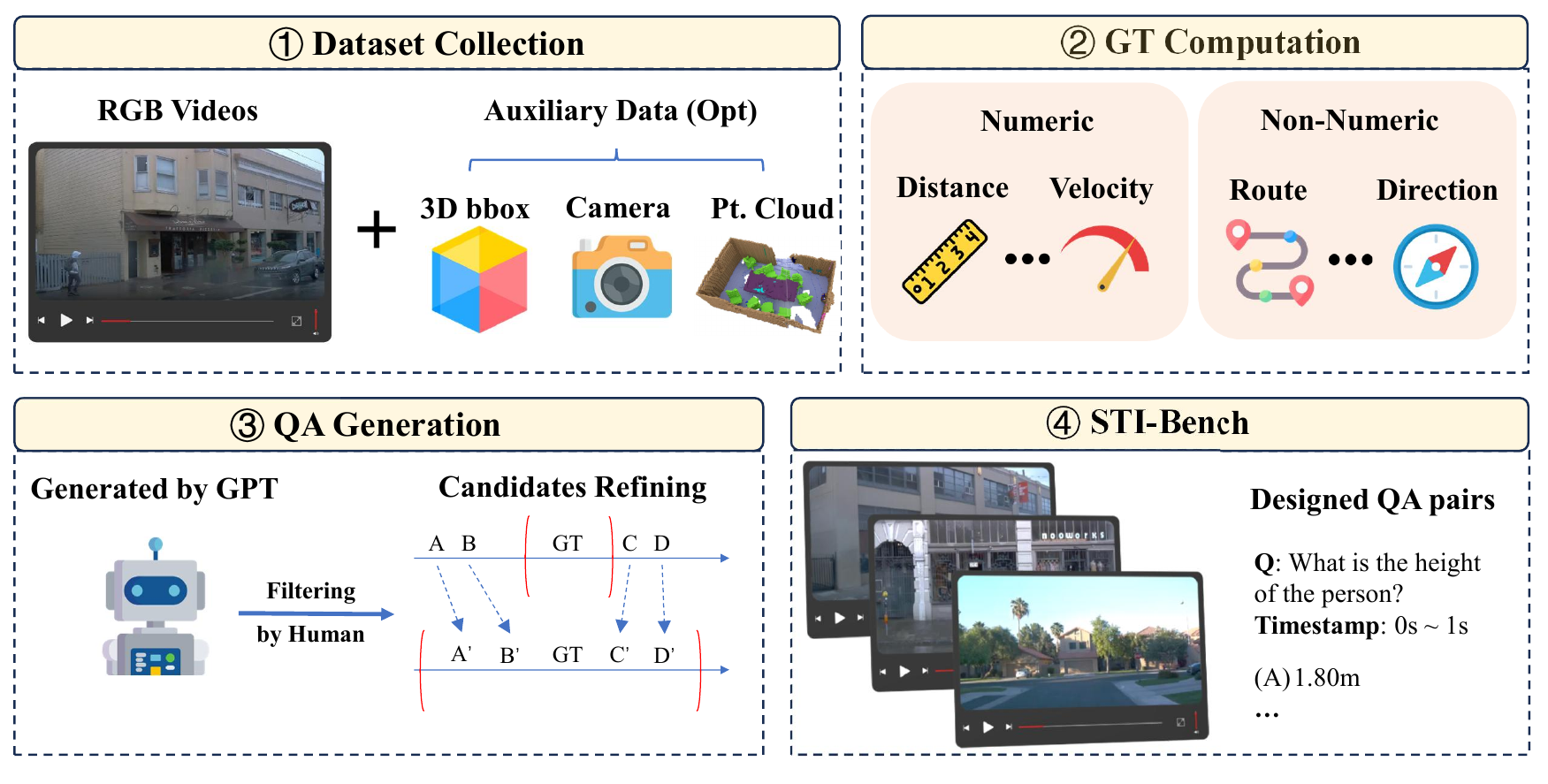}
    \caption{\textbf{Benchmark curation pipeline.} The pipeline first aggregates multi-scene RGB datasets that contain 3D bounding box annotations, camera parameters, and point cloud data, which serve as the basis for computing ground truth. From these datasets, we extract numerical ground truth such as distance and velocity, as well as textual descriptions of trajectories and directions. Subsequently, we leverage GPT to assist in generating QA pairs and design a website for rigorous human verification and filtering.}
    \label{fig: pipeline}
\end{figure*}

In this section, we present the detailed design and construction of STI-Bench. The construction pipeline is depicted in Figure \ref{fig: pipeline}.

\subsection{Task Definition}
We propose eight tasks examining distinct aspects of MLLMs' spatial-temporal understanding, divided into Static Understanding and Dynamic Understanding.

\noindent \textbf{Static Understanding}
\begin{enumerate}[label=\alph*.]
    \item \textbf{Dimensional Measurement.} Estimates object geometric size and distances, requiring transformation from 2D pixel observations to physical measurements.\\
    \textit{Example:} "What is the height of this box?"
    
    \item \textbf{Spatial Relation.} Identifies spatial relationships among objects or between camera and objects, testing relative positioning understanding across viewpoints.\\
    \textit{Example:} "Is the chair on the left or right of the table?"
    
    \item \textbf{3D Video Grounding.} Retrieves object's 3D bounding box given semantic descriptions, requiring alignment of linguistic and visual features.\\
    \textit{Example:} "Locate the 3D bounding box of the red suitcase near the bed."
\end{enumerate}

\noindent \textbf{Dynamic Understanding}
\begin{enumerate}[label=\alph*., start=4]
    \item \textbf{Displacement and Path Length.} Measures travel distance between time points, requiring motion tracking across frames.\\
    \textit{Example:} "How far has the car traveled from 1s to 18s?"
    
    \item \textbf{Speed and Acceleration.} Computes motion parameters by integrating spatial displacement with time intervals.\\
    \textit{Example:} "What is the average speed of the camera?"
    
    \item \textbf{Ego-Centric Orientation.} Examines camera's azimuth orientation changes, requiring understanding of rotation representations.\\
    \textit{Example:} "How many degrees does the camera's horizontal orientation shift?"
    
    \item \textbf{Trajectory Description.} Describes motion paths throughout the video, testing ability to abstract spatial motion into language.\\
    \textit{Example:} "Summarize the camera trajectory, including distances moved and turns made."
    
    \item \textbf{Pose Estimation.} Estimates camera pose at specified time points using RGB data, requiring visual odometry capabilities.\\
    \textit{Example:} "Given the initial pose, what is the camera's pose at the requested time?"
\end{enumerate}

These tasks collectively evaluate comprehensive spatial-temporal intelligence across different scales, requiring fundamental 3D spatial reasoning, physical common sense, and cross-modal information integration over time.

\subsection{Benchmark Construction}

\paragraph{Data Collection.}
To encompass a broad spectrum of real-world environments, STI-Bench covers three major scenarios: \textit{Desktop}, \textit{Indoor}, and \textit{Outdoor}. Accordingly, we draw from three publicly available datasets—\textbf{Waymo}~\cite{Sun_2020_CVPR} for autonomous driving, \textbf{ScanNet}\cite{dai2017scannet} for indoor 3D scene reconstruction, and \textbf{Omni6DPose}\cite{zhang2024omni6dpose} for desktop-scale 6D object pose estimation. These datasets provide frame-by-frame camera intrinsic and extrinsic parameters, as well as point clouds for each object, which we map to two-dimensional bounding boxes in each frame.

\paragraph{Automatic QA Pair Generation.}
We used MLLMs to produce detailed semantic descriptions for each object, such as “A beige minivan with a roof rack,” “A refrigerator with emoji magnets, photos, and a to-do list,” or “A red backpack on a brown leather sofa.” Next, leveraging the frame-by-frame annotations, we computed the ground-truth information required for each task. We then provided the ground-truth data, object descriptions, and task-specific QA requirements to MLLMs to generate a diverse set of questions and challenging answer options.

\paragraph{Human Quality Control.}
During QA pair generation, several issues arose: LLM-generated descriptions could be inaccurate or fail to uniquely identify the target object; some questions and options remained unreasonable or incorrect, even with detailed guidelines; and in certain cases, the video alone did not provide sufficient information (e.g., when the camera was occluded but lidar data were available).
To address these challenges, we developed a website for multiple rounds of manual filtering and sampling-based review, ensuring high-quality questions. We also randomly shuffled the answer options to enhance evaluation robustness. Ultimately, we curated more than 2,000 high-quality QA pairs from over 300 videos. Details are shown in Figure~\ref{fig:statistic}.

\paragraph{Fine-Grained Adjustment.}
To ensure task diversity and challenge, each question is equipped with carefully constructed distractor options that are both significantly different from the correct answer while remaining within a reasonable range. Considering the varying precision requirements across different scenarios—millimeter-level for desktop applications, centimeter-to-decimeter-level for indoor environments, and decimeter-to-meter-level for outdoor scenes—we applied scenario-specific scaling factors to generate distractors and employed logarithmic sampling techniques to distribute numerical differences more evenly, thereby enhancing the robustness of our evaluation.

Specifically, given a scene type $S \in \{\text{Desktop}, \text{Indoor}, \text{Outdoor}\}$, a correct answer $A_{correct}$, and four initial distractor options $\{A_{d1}, A_{d2}, A_{d3}, A_{d4}\}$, our fine-grained adjustment process consists of the following steps:

1. We determine the error range $[E_{min}, E_{max}]$ based on the scene type $S$:
   - Desktop: $[0.5\text{ cm}, 5\text{ cm}]$
   - Indoor: $[5\text{ cm}, 50\text{ cm}]$
   - Outdoor: $[0.5\text{ m}, 5\text{ m}]$

2. We logarithmically sample an error value $e$ from the corresponding error range:
   $$e = E_{min} \cdot (E_{max}/E_{min})^u$$
   where $u \sim \mathcal{U}(0,1)$ is a uniform random number.

3. We adjust each distractor option $A_{di}$ to $A_{di}'$ using a weighted average:
   $$A_{di}' = (1-w) \cdot A_{di} + w \cdot A_{correct}$$
   where the weight $w$ is identical for all distractors and is determined by solving:
   $$\min_{i \in \{1,2,3,4\}} \|A_{di}' - A_{correct}\| = e$$
   
   This yields:
   $$w = 1 - \frac{e}{\min_{i \in \{1,2,3,4\}} \|A_{di} - A_{correct}\|}$$

This approach ensures that the minimum distance between any adjusted distractor and the correct answer equals the sampled error value $e$, while other distractors maintain relatively larger distances but are adjusted using the same weight factor.

\begin{table*}[t]
\centering
\setlength\tabcolsep{3pt}
\resizebox{\textwidth}{!}{
\begin{tabularx}{\textwidth}{rcc>{\centering\arraybackslash}X*{7}{>{\centering\arraybackslash}X}}
\Xhline{1pt}
\multirow{3}{*}{\textbf{Methods}} & \multirow{3}{*}{\textbf{Rank}} & \multirow{3}{*}{\textbf{Avg.}} &
\multicolumn{3}{c}{\textbf{Static Understanding}} &
\multicolumn{5}{c}{\textbf{Dynamic Understanding}} \\
\cmidrule(lr){4-6} \cmidrule(lr){7-11}
& & &
\multirow{2}{*}{\makecell{Dim.\\Meas.}} &
\multirow{2}{*}{\makecell{Spatial\\Relation}} &
\multirow{2}{*}{\makecell{3D Video\\Grounding}} &
\multirow{2}{*}{\makecell{Disp.\\\& P.L.}} &
\multirow{2}{*}{\makecell{Speed\\\& Acc.}} &
\multirow{2}{*}{\makecell{Ego\\Orient.}} &
\multirow{2}{*}{\makecell{Traj.\\Desc.}} &
\multirow{2}{*}{\makecell{Pose\\Est.}} \\
& & & & & & & & & & \\ \hline
\rowcolor{lightblue!20} \multicolumn{11}{l}{\textcolor{black}{\textit{Proprietary Models (API)}}} \\[-0.5ex]
GPT-4o\cite{gpt4o}              & 8 & 34.8 & 27.1 & \cellcolor{light-orange}{51.8} & 29.0 & 23.2 & 35.4 & 33.7 & 32.0 & 53.6 \\
Gemini-2.0-Flash\cite{team2024gemini} & 4 & 38.7 & 31.9 & 50.0 & 31.8 & 27.7 & 32.1 & 10.8 & 38.5 & \cellcolor{light-orange}{61.3} \\
Claude-3.7-Sonnet\cite{claude}   & 3 & 40.5 & 29.8 & 45.5 & 35.7 & \cellcolor{light-orange}{28.9} & \cellcolor{orange}{38.8} & 40.0 & 47.4 & \cellcolor{orange}{62.6} \\
Gemini-2.5-Pro\cite{gemini25}    & 1 & \cellcolor{orange}{41.4} & \cellcolor{orange}{38.7} & \cellcolor{orange}{53.8} & \cellcolor{light-orange}{36.9} & \cellcolor{orange}{33.9} & 33.1 & \cellcolor{orange}{52.5} & \cellcolor{light-orange}{47.4} & 50.4 \\
\hline
\rowcolor{lightblue!20} \multicolumn{11}{l}{\textcolor{black}{\textit{Open-source Models}}} \\[-0.5ex]
MiniCPM-V-2.6\cite{yao2024minicpm}  & 10 & 26.9 & 27.7 & 44.5 & 29.0 & 19.0 & 25.7 & 7.0  & 30.8 & 35.6 \\
VideoChat-R1\cite{videochatr1}     & 9  & 32.8 & 23.2 & 47.3 & 31.5 & 22.4 & 31.1 & 26.0 & 47.9 & 48.3 \\
VideoLLaMA3-7B\cite{videollama3}   & 7  & 35.2 & 29.4 & 48.6 & 36.1 & 21.5 & 36.7 & 23.2 & \cellcolor{orange}{54.6} & 48.1 \\
VideoChat-Flash\cite{li2024videochat} & 6 & 36.3 & \cellcolor{light-orange}{33.6} & 51.4 & 33.1 & 27.1 & 32.3 & 22.2 & \cellcolor{light-orange}{54.2} & 51.4 \\
InternVL2.5-78B\cite{internvl}    & 5  & 38.5 & 29.9 & \cellcolor{light-orange}{52.8} & 31.6 & 24.9 & 37.2 & \cellcolor{light-orange}{49.2} & 43.6 & 53.6 \\
Qwen2.5-VL-72B\cite{qwen25vl}     & 2  & \cellcolor{light-orange}{40.7} & 31.5 & 47.6 & \cellcolor{orange}{39.1} & 25.1 & \cellcolor{light-orange}{38.4} & 43.8 & 51.3 & 60.6 \\
\Xhline{1pt}
\end{tabularx}}
\caption{\textbf{Evaluation on STI-Bench.} \colorbox{orange}{Orange} = best; \colorbox{light-orange}{Light Orange} = second best.\textbf{Note: The random baseline for all tasks is 20\%.}}
\label{tab:results}
\end{table*}

\begin{table}[htbp]
\centering
\resizebox{0.9\columnwidth}{!}{%
\renewcommand{\arraystretch}{1.15}
\setlength{\tabcolsep}{2mm}
\begin{tabular}{lcccc}
    \toprule
    \textbf{Model} & \textbf{Outdoor} & \textbf{Indoor} & \textbf{Desktop} & \textbf{Overall} \\
    \midrule
    MiniCPM-V-2.6\cite{yao2024minicpm}        & 27.9 & 26.1 & 26.6 & 26.9 \\
    VideoChat-R1\_7B\cite{videochatr1}        & 41.1 & 28.0 & 27.1 & 32.8 \\
    GPT-4o\cite{gpt4o}                        & 41.4 & 33.1 & 27.3 & 35.2 \\
    VideoLLaMA3-7B\cite{videollama3}          & 41.2 & 31.9 & 30.5 & 35.2 \\
    Gemini-2.0-Flash\cite{team2024gemini}     & 39.1 & 35.9 & 30.5 & 36.0 \\
    VideoChat-Flash\cite{li2024videochat}     & 38.6 & 36.7 & 31.3 & 36.3 \\
    InternVL2.5-78B\cite{internvl}            & 46.1 & 34.5 & 32.1 & 38.5 \\
    Claude-3.7-Sonnet\cite{claude}            & 47.2 & \cellcolor{orange}{38.2} & 32.3 & 40.5 \\
    Qwen2.5-VL-72B\cite{qwen25vl}             & \cellcolor{orange}{50.6} & 35.0 & \cellcolor{light-orange}{33.5} & \cellcolor{light-orange}{40.7} \\
    Gemini-2.5-Pro\cite{gemini25}             & \cellcolor{light-orange}{48.7} & \cellcolor{light-orange}{37.1} & \cellcolor{orange}{35.8} & \cellcolor{orange}{41.4} \\
    \bottomrule
\end{tabular}}
\caption{\textbf{Evaluation results across different scenes.}}
\label{tab:results_categories}
\end{table}

\begin{figure*}[htbp]
    \centering
    \includegraphics[width=0.9\linewidth]{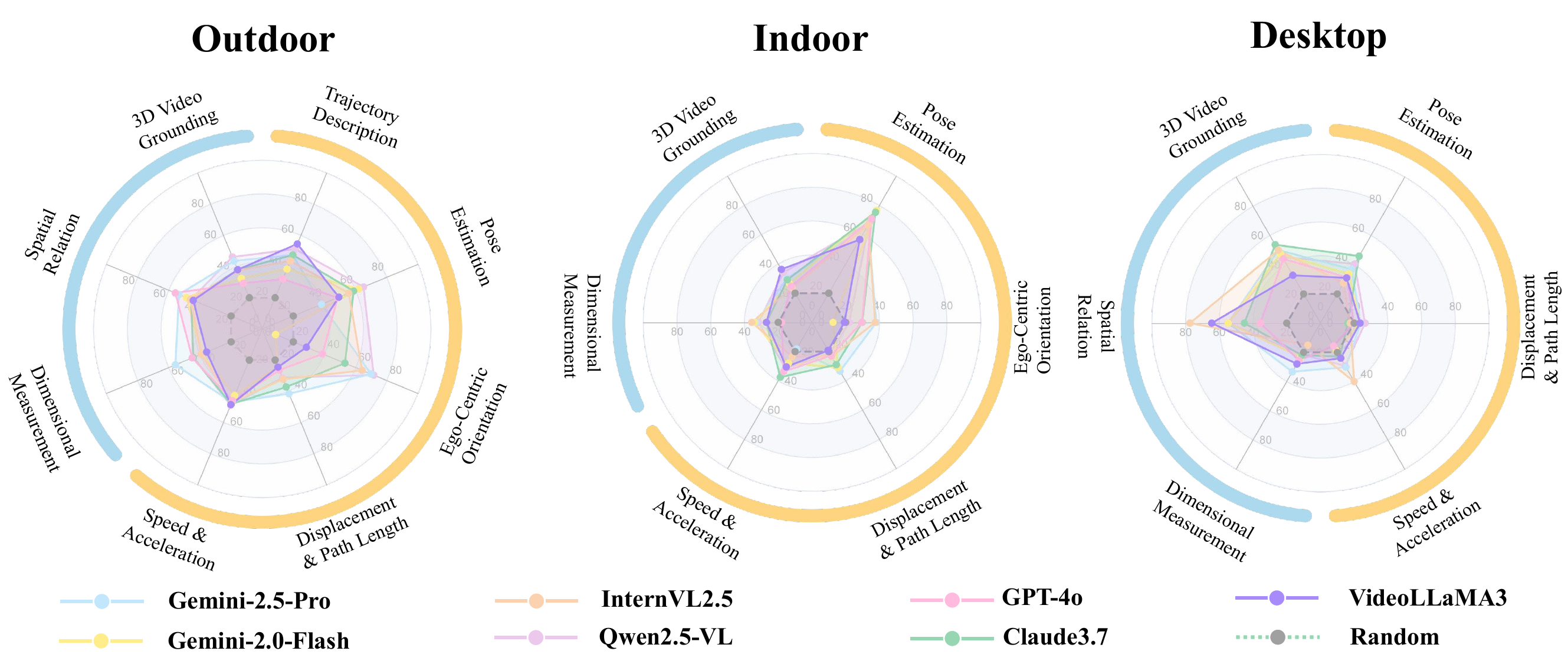   }
    \caption{\textbf{Evaluation results across different scenes and tasks.}}
    \label{fig:three_subfigures}
\end{figure*}

\section{Experiments}
\subsection{Settings}
We conduct a thorough evaluation of leading MLLMs from diverse model families, focusing on both proprietary and open-source solutions. Specifically, we assess four proprietary models: GPT-4o\cite{gpt4o}, Gemini-2.0-Flash\cite{team2024gemini}, Gemini-2.5-Pro\cite{team2024gemini}, and Claude-3.7-Sonnet\cite{claude}, as well as several representative open-source MLLMs that have undergone specialized video-related training, including Qwen2.5-VL-72B\cite{qwen25vl}, InternVL2.5-78B\cite{internvl}, MiniCPM-V-2.6\cite{yao2024minicpm}, VideoChat-Flash\cite{li2024videochat}, VideoChat-R1\cite{videochatr1} and VideoLLaMA3-7B\cite{videollama3}.  

Considering the stability of open-source models and the API limitations of proprietary models, we uniformly sample 30 frames from the video for each record and explicitly indicate the sampling FPS within the prompt. An exception is made for Claude-3.7-Sonnet, for which only 20 frames are sampled due to its API constraints. Our benchmark tasks are presented in a multiple-choice format with five possible answers, hence a random guess baseline yields 20\% accuracy. We measure each model's accuracy by directly comparing the model's selected answer with the ground truth.

\begin{figure*}[t]
    \centering
    \includegraphics[width=0.9\textwidth]{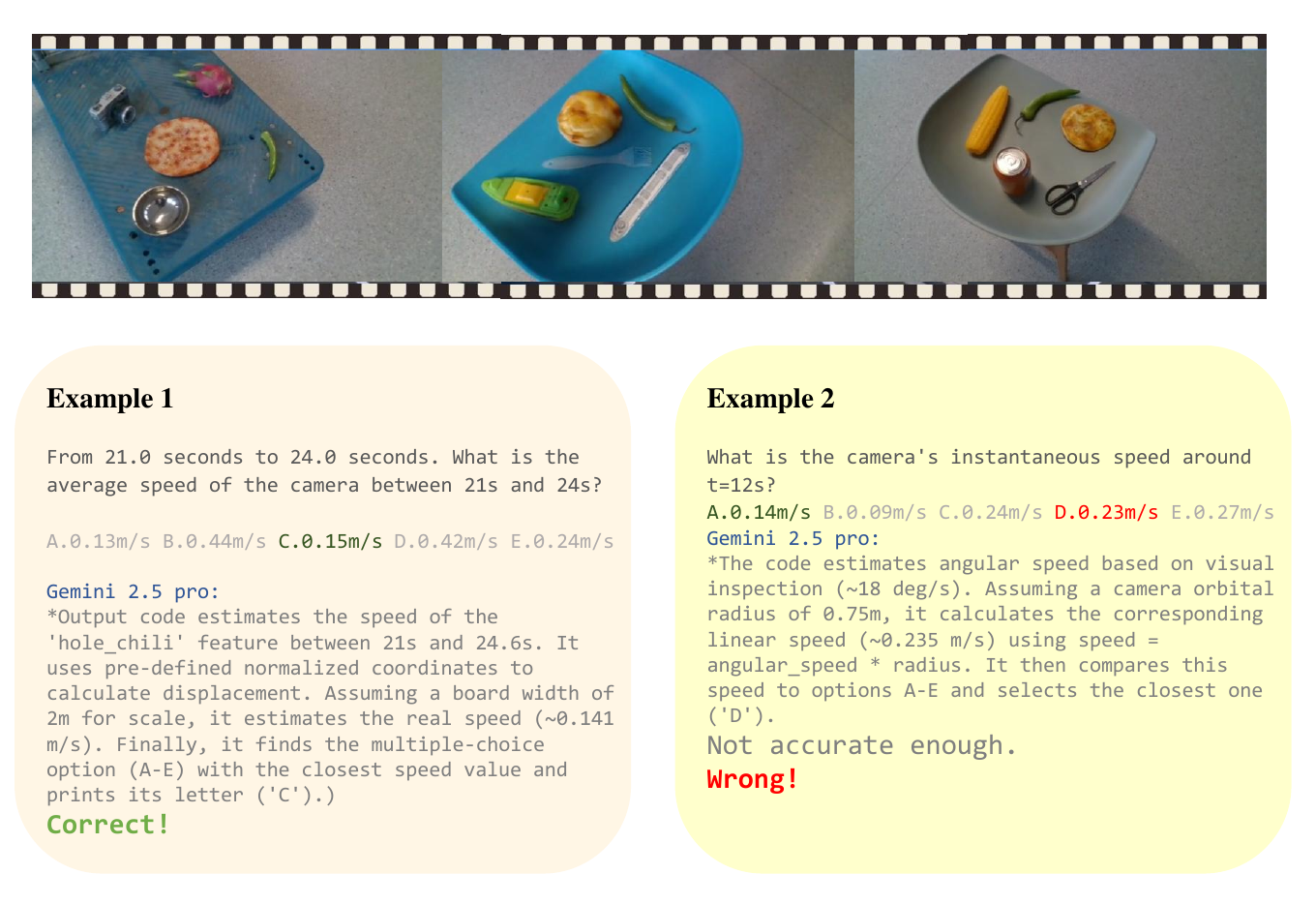}
    \caption{The simplified thought process examples of Gemini 2.5 Pro.}
    \label{fig:reason}
\end{figure*}

\subsection{Main Results}
As shown in Table~\ref{tab:results} and Table~\ref{tab:results_categories}, we present a comprehensive evaluation of various MLLMs on STI-Bench.
Gemini-2.5-Pro achieves the highest average accuracy of 41.4\%, closely followed by Qwen2.5-VL-72B at 40.7\% among open-source models.
While these results significantly exceed the random guess baseline (20\%), they highlight substantial room for improvement in spatial-temporal understanding.

Performance across different scene types shows varied patterns.
Qwen2.5-VL-72B demonstrates strong performance in outdoor scenarios (50.6\%), while Gemini-2.5-Pro performs well outdoors (48.7\%) and leads in desktop scenes (35.8\%).
In indoor environments, Claude-3.7-Sonnet (38.2\%) and Gemini-2.5-Pro (37.1\%) lead.
These variations suggest different model specializations, possibly influenced by training data distribution.

Task-specific performance reveals particularly challenging areas.
Models generally struggle with precise quantitative estimation tasks.
The best performance on Dimensional Measurement is achieved by Gemini-2.5-Pro (38.7\%), and on Displacement \& Path Length also by Gemini-2.5-Pro (33.9\%).
In contrast, models demonstrate stronger capabilities in Pose Estimation (best: 62.7\% by Claude-3.7-Sonnet) and Spatial Relation tasks (best: 53.8\% by Gemini-2.5-Pro).

Among open-source models, Qwen2.5-VL-72B stands out with 40.7\% average accuracy, demonstrating highly competitive performance.
Other open-source models like InternVL2.5-78B (38.5\%) also show promise, though smaller models like MiniCPM-V-2.6 (26.9\%) lag behind their larger counterparts.

Even the best-performing model, Gemini-2.5-Pro, achieves only 41.4\% average accuracy on our benchmark.
These results indicate that current MLLMs, despite impressive general visual understanding capabilities, still require significant advancements in precise spatial-temporal intelligence for embodied tasks.

\subsection{Experimental Analysis}
Given that Gemini-2.5-Pro exhibits strong multi-modal reasoning capabilities and achieves the top overall performance (41.4\% average accuracy), we select it as a representative for in-depth analysis.

Performance varies across scene types: it performs strongest in outdoor scenarios (48.7\%), followed by indoor (37.1\%) and desktop environments (35.8\%). This disparity might suggest that the model's training data is better attuned to outdoor and larger-scale understanding.

Examining task-specific performance, we observe its strongest capabilities in tasks involving relative understanding and object state. It achieves high scores in Spatial Relation (53.8\%) and Ego-Centric Orientation (52.5\%), followed by Pose Estimation (50.4\%). While tasks requiring precise quantitative estimation show lower absolute scores, Gemini-2.5-Pro performs competitively and often leads in these areas, achieving top scores in Dimensional Measurement (38.7\%) and Displacement \& Path Length (33.9\%).

\begin{figure}[t]
    \centering
    \includegraphics[width=\columnwidth]{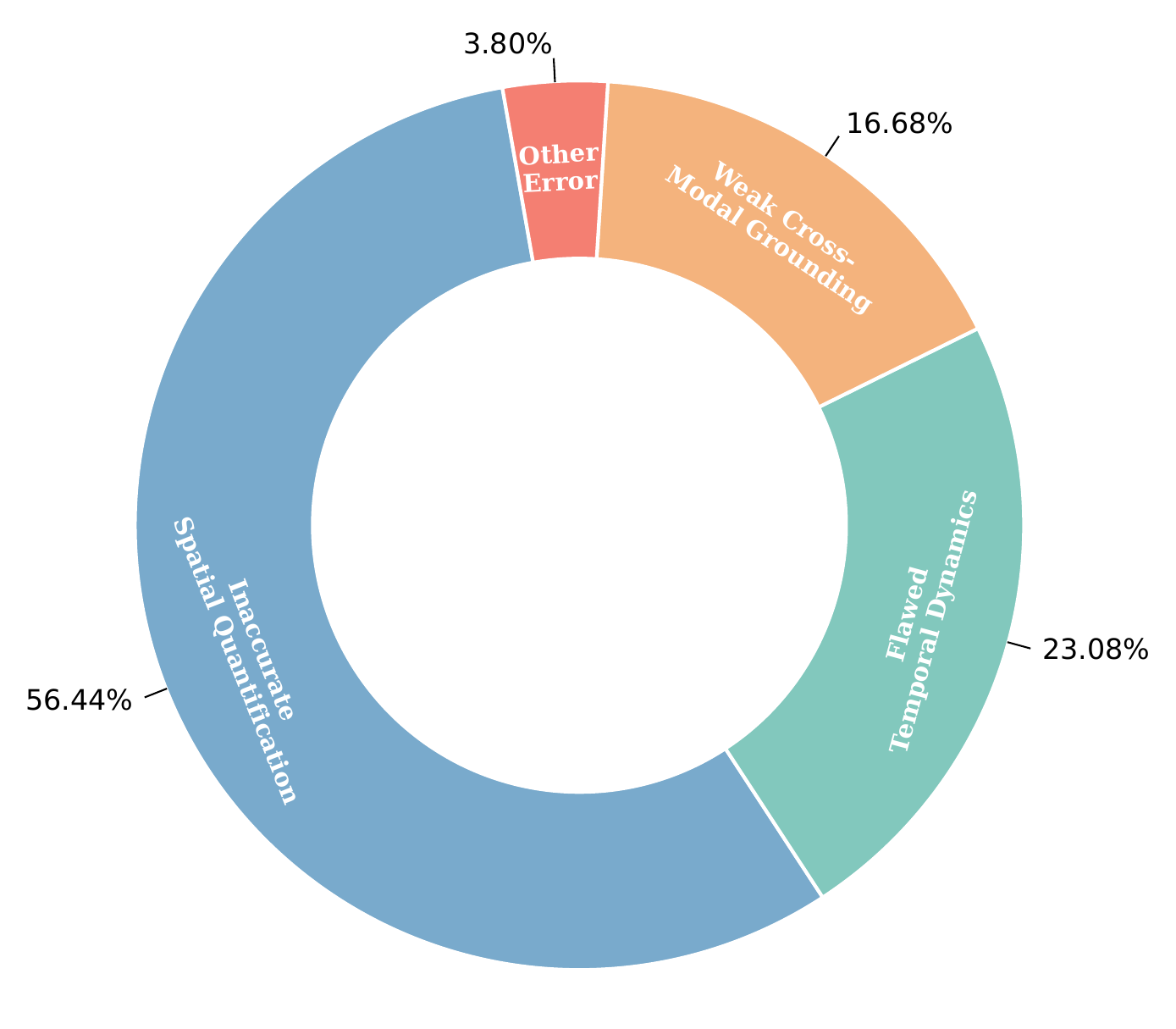}
    \caption{Distribution of error categories in Gemini-2.5-Pro across our sampled error cases.}
    \label{fig:error}
\end{figure}

By leveraging the model's reasoning process and uniformly sampling approximately 200 error records across each task type and scenario, we categorize errors into three representative patterns. Figure~\ref{fig:error} shows the distribution of error categories.

\paragraph{Inaccurate Spatial Quantification}
The model faces significant challenges in accurately estimating static spatial properties and relationships from visual inputs. These difficulties manifest when estimating object dimensions, distances between objects or between camera and objects, and absolute 3D coordinates at specific time points. These errors stem from a lack of clear visual size references, difficulty distinguishing between numerically close options, and the inherent challenges of inferring metric scale from 2D pixels and estimating depth with monocular cameras. Such limitations directly impact performance in dimensional measurement, spatial relation, and 3D video grounding tasks.

\paragraph{Flawed Temporal Dynamics Understanding}
The model struggles to perceive, track, and interpret cross-frame information that changes over time, such as motion and its dynamics. This results in erroneous calculations or descriptions of displacement, path length, speed, acceleration, directional changes, and overall trajectory shapes. The model particularly struggles with relative motion (distinguishing object motion from camera motion), a problem exacerbated by sparse temporal sampling. These difficulties arise from challenges in integrating information across frames, lack of internal models for physics/kinematics, inability to separate ego-motion from object motion, and information loss due to sparse sampling. These issues manifest in tasks involving displacement and path length, speed and acceleration, ego-centric orientation, trajectory description, and pose estimation.

\paragraph{Weak Cross-Modal Grounding and Integration}
The model fails to properly connect textual queries/instructions with relevant spatial-temporal visual content, or to integrate provided non-visual data (such as initial poses) with visual information. This includes misinterpreting temporal constraints (like "from 1s to 18s," "at the end," "the moment of last co-occurrence"), failing to correctly utilize provided initial conditions (e.g., initial camera pose in pose estimation tasks), and incorrectly associating structured data (coordinates, timestamps) with visual elements. These errors stem from deficiencies in parsing structured/natural language instructions and difficulty integrating information from different modalities into a unified reasoning process. This affects all tasks that rely on specific instructions or initial data.

These error patterns highlight that, despite Gemini-2.5-Pro's strong performance relative to other models, it still faces significant challenges in precise spatial-temporal understanding. Its limitations in quantitative estimation and complex spatial-temporal reasoning indicate that current MLLMs remain far from achieving the reliability required for embodied AI or autonomous driving applications.

\section{Conclusion}



We introduced STI-Bench, a comprehensive benchmark to assess MLLMs’ spatial-temporal understanding through over 300 real-world videos and 2,000 QA pairs of robot outdoor, indoor, and desktop scenarios, which reveals significant limitations in current MLLMs' spatial-temporal understanding capabilities, with top-performing models like Gemini-2.5-Pro achieving around 41.4\% average accuracy. Models particularly struggle with precise quantitative tasks like dimensional measurement. Our analysis identifies three key weaknesses: inaccurate spatial quantification, flawed temporal dynamics understanding, and weak cross-modal integration. These findings emphasize the substantial gap between current capabilities and the reliability needed for embodied AI and autonomous driving applications. STI-Bench provides a valuable framework for evaluating and improving MLLMs' ability to understand the physical world—essential for developing the next generation of embodied intelligent systems.

\section*{Acknowledgement}

This work is funded by NSFC-62306046. We thank the data annotation team of MolarData for their support, as well as the volunteer contributors Yuxin Liu, Junjie Ruan, Xucheng Liao, Zixuan Huang, Tianrui Wan, Qingcheng Wei, Yujie Yao, Shangyang Dong, Zhuofan Zeng and Yiming Li for their valuable work as human annotators and evaluators.

{
    \small
    \bibliographystyle{ieeenat_fullname}
    \bibliography{main}
}

\clearpage

\appendix 


\renewcommand{\thesection}{\Alph{section}} 
\renewcommand{\thesubsection}{\thesection.\arabic{subsection}} 
\renewcommand{\thesubsubsection}{\thesubsection.\arabic{subsubsection}} 

\section{Details for Benchmark Construction}

\subsection{Dataset Interface} 
\label{sec:interface}
When searching for datasets, we found that apart from RGB frame data, other data formats are inconsistent. For example, we noticed that it does not provide 3D bounding box information, so we could only use it to compute camera displacement, velocity, and other related physical quantities. In contrast, Waymo includes such information, which is why we aim to align different datasets into a unified format. Therefore, we unify different types of data by converting them into frame-by-frame instance-level 3D point clouds or bounding box annotations, along with per-frame camera parameters, enabling us to obtain all the required information and process it in a consistent manner.

\subsection{Ground Truth Computation}
\label{sec:gt_computation}

In STI-Bench, the ground truth annotations are derived from multi-modal data, including frame-by-frame camera intrinsics/extrinsics and precise 3D point cloud annotations. We utilize three publicly available datasets (Waymo, ScanNet, Omni6DPose) to cover different scenarios (outdoor, indoor, desktop). Below, we provide eight benchmark tasks with unified notations and formulas.

\subsubsection{Static Understanding}

\paragraph{Dimensional Measurement}
Let $l_x$, $l_y$, $l_z$ denote the dimensions (length, width, height) of an object along the $x$, $y$, and $z$ axes:
\begin{equation}
\label{eq:lxlylz}
\begin{aligned}
l_x &= x_{\max} - x_{\min},\\
l_y &= y_{\max} - y_{\min},\\
l_z &= z_{\max} - z_{\min}.
\end{aligned}
\end{equation}
Here, $l_x, l_y, l_z$ represent the object size along each coordinate axis.

If we need the distance between two objects (or between the camera and an object), let $d_{12}$ be the Euclidean distance between their center points:
\begin{equation}
\label{eq:d12}
d_{12} = \sqrt{(x_2 - x_1)^2 + (y_2 - y_1)^2 + (z_2 - z_1)^2}.
\end{equation}
Here, $(x_1, y_1, z_1)$ and $(x_2, y_2, z_2)$ are the center coordinates of the two objects.

\paragraph{Spatial Relation}
When the difference along one coordinate axis is significantly larger than along others, the sign of that difference determines the spatial relation:
\begin{equation}
\label{eq:spatialrelation}
\begin{aligned}
r_{xy} &= \mathrm{sign}(x_A - x_B),\\
r_{yz} &= \mathrm{sign}(y_A - y_B),\\
r_{zx} &= \mathrm{sign}(z_A - z_B).
\end{aligned}
\end{equation}
Here, $r_{xy}, r_{yz}, r_{zx}$ indicate relative positioning along each axis (e.g., front/back, left/right, above/below). We choose the axis with the greatest difference to label the dominant relation.

\paragraph{3D Video Grounding}
For frame $t$ in the camera coordinate system, the 3D bounding box of an object can be described with dimensions, center position, and optional rotations:
\begin{equation}
\label{eq:bboxt}
\text{BBox}_{t} = (\,l_t,\, w_t,\, h_t,\, x_t,\, y_t,\, z_t,\, \text{yaw}_t,\, \text{pitch}_t,\, \text{roll}_t\,).
\end{equation}
Here, $(l_t, w_t, h_t)$ are the object dimensions, $(x_t, y_t, z_t)$ is the center position, and $(\text{yaw}_t, \text{pitch}_t, \text{roll}_t)$ are optional rotation angles if available.

\subsubsection{Dynamic Understanding}

\paragraph{Pose Estimation}
Given the camera's initial pose $(p_0, o_0)$, the pose $(p_t, o_t)$ at time $t$ can be obtained using the extrinsic-derived matrices $R_t$ (rotation) and $T_t$ (translation):
\begin{equation}
\label{eq:pose}
\begin{aligned}
p_t &= R_t \, p_0 + T_t,\\
o_t &= o_0 + \Delta o_t.
\end{aligned}
\end{equation}
Here, $p_t$ is the position, $o_t$ is the orientation.

\paragraph{Displacement and Path Length}
Let $p_i = (x_i, y_i, z_i)$ be the position at time $i$. The displacement $d_{0n}$ and path length $L_{\text{traj}}$ are computed as:
\begin{equation}
\label{eq:displacement}
d_{0n} = \sqrt{(x_n - x_0)^2 + (y_n - y_0)^2 + (z_n - z_0)^2},
\end{equation}
\begin{equation}
\label{eq:pathlength}
L_{\text{traj}} = \sum_{i=1}^{n} \sqrt{(x_i - x_{i-1})^2 + (y_i - y_{i-1})^2 + (z_i - z_{i-1})^2}.
\end{equation}
Here, $d_{0n}$ is the straight-line distance from the initial to the final position; $L_{\text{traj}}$ sums consecutive segment lengths for the entire path.

\paragraph{Speed and Acceleration}
Let $\Delta t$ be the time interval between consecutive frames. Then the speed $v_i$ and acceleration $a_i$ are:
\begin{equation}
\label{eq:speedacc}
\begin{aligned}
v_i &= \frac{d_i}{\Delta t},\\
a_i &= \frac{v_i - v_{i-1}}{\Delta t}.
\end{aligned}
\end{equation}
Here, $d_i$ is the displacement between adjacent frames, $v_i$ is the speed at time $i$, and $a_i$ is the acceleration.

\paragraph{Ego-Centric Orientation}
If $\theta_t$ denotes the camera orientation (azimuth) at time $t$, then the orientation change $\Delta \theta_t$ is:
\begin{equation}
\label{eq:orientation}
\Delta \theta_t = \theta_t - \theta_0.
\end{equation}
This indicates how much the camera has rotated relative to its initial azimuth.

\paragraph{Trajectory Description}
We apply the Ramer-Douglas-Peucker (RDP) algorithm to simplify the sequence of positions into key line segments. The resulting polyline is described in a piecewise manner (e.g., ``go straight for 30m, turn left 85°, then go straight for 20m, ...''), providing a concise representation of complex motion trajectories.

\subsection{Human Involved Quality Control}
\label{sec:quality}

Despite leveraging automated methods for generating large quantities of question-answer pairs, several critical issues necessitated human intervention for quality assurance. These included inaccuracies or ambiguities in automatically generated object descriptions, logical inconsistencies or irrelevant content within questions and options, and insufficient visual information in video data due to occlusions or limited resolution. To address these challenges, we established a dedicated online platform for systematic multi-round manual review and random sampling checks. Human reviewers evaluated each QA pair based on several criteria: clarity and accuracy of object descriptions, logical coherence between questions and answer options, sufficiency of visual information to accurately respond to questions, and effectiveness of distractors in posing genuine challenges. After passing the rigorous review process, answer options were randomized to prevent bias arising from fixed option ordering. Ultimately, approximately 20\% of all generated QA pairs were retained. This meticulous procedure enabled us to select over 2,000 high-quality QA pairs from more than 300 video sequences, significantly enhancing the robustness and reliability of the STI-Bench benchmark dataset.

\section{Adjustment Details}

To ensure option diversity under different positioning accuracies,
we generate refined distractors for each scene
$S\in\{\mathrm{desktop},\mathrm{indoor},\mathrm{outdoor}\}$.
Given the ground-truth answer $A_{\text{c}}$
and four initial distractors $\{A_{d,1},A_{d,2},A_{d,3},A_{d,4}\}$,
the adjustment consists of three steps.

\paragraph{Step\,1: Scene-dependent error interval.}
Define $[E_{\min},E_{\max}]$ as
\[
[E_{\min},E_{\max}] =
\begin{cases}
[0.5\,\mathrm{cm},\;5\,\mathrm{cm}], & S=\mathrm{desktop},\\[2pt]
[5\,\mathrm{cm},\;50\,\mathrm{cm}], & S=\mathrm{indoor},\\[2pt]
[0.5\,\mathrm{m},\;5\,\mathrm{m}], & S=\mathrm{outdoor}.
\end{cases}
\]

\paragraph{Step\,2: Log-uniform sampling of target error.}
Sample $u\sim\mathcal U(0,1)$ and compute
\begin{equation}\label{eq:log-sample}
e = E_{\min}\Bigl(\tfrac{E_{\max}}{E_{\min}}\Bigr)^{u}.
\end{equation}

\paragraph{Step\,3: Joint adjustment of distractors.}
All distractors are shifted towards $A_{\text{c}}$
with a common weight $w$:
\begin{equation}\label{eq:adjust}
A_{d,i}' = (1-w)A_{d,i}+wA_{\text{c}},\qquad i=1,\dots,4.
\end{equation}
The weight $w$ is chosen such that the closest adjusted distractor
attains the target error $e$:
\begin{equation}\label{eq:w-solve}
w = 1-\frac{e}{\displaystyle\min_{i}\bigl\lVert A_{d,i}-A_{\text{c}}\bigr\rVert}.
\end{equation}

By sharing the same $w$, the minimum
gap between any distractor and the correct answer is exactly $e$,
while others remain further apart, thereby preserving option dispersion.

\section{Detailed Results}

\begin{figure*}[h!]
    \centering
    \includegraphics[width=\textwidth]{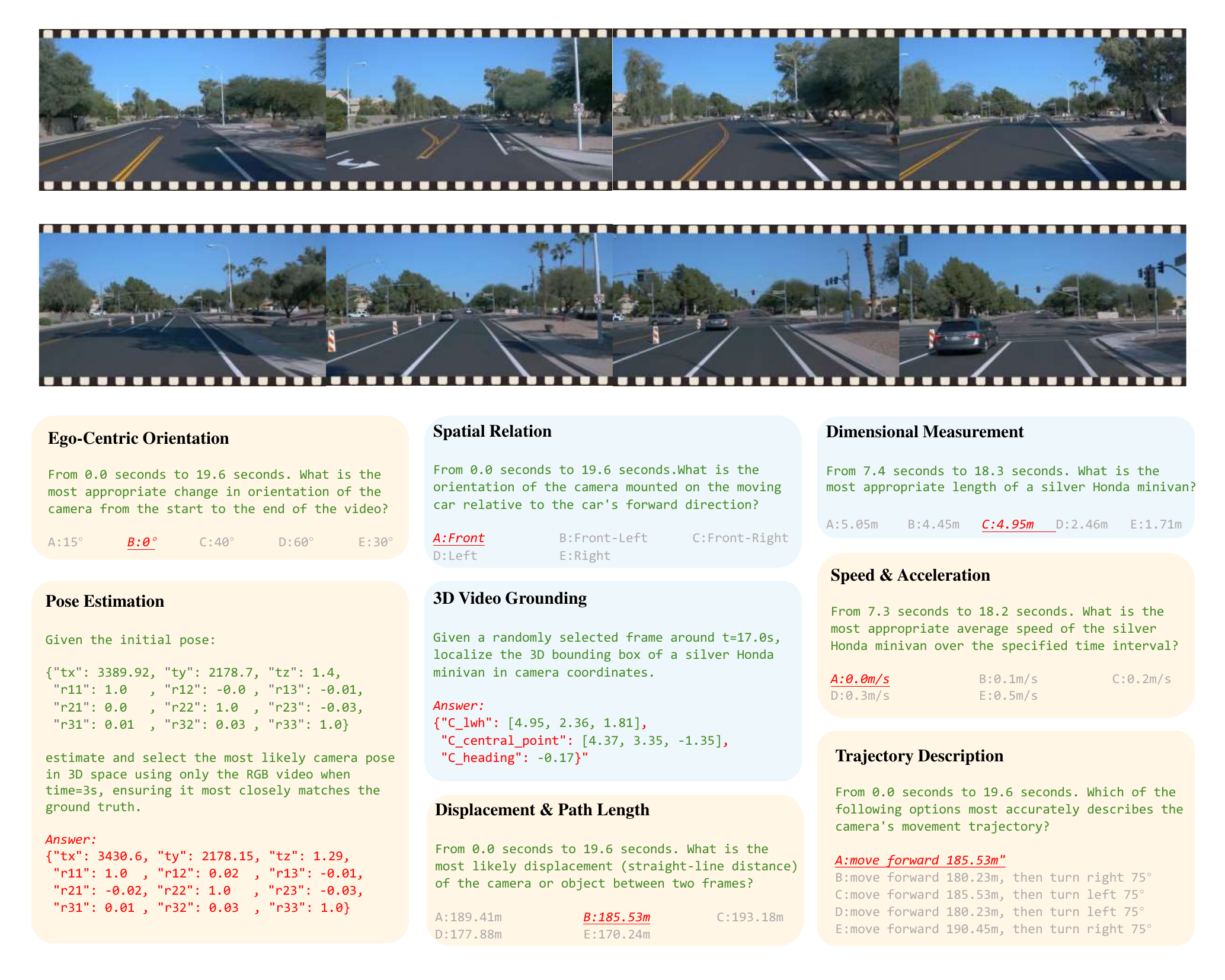}
    \caption{\textbf{STI-Bench Examples (Outdoor)} }
    \label{fig:outdoor_example}
\end{figure*}

\begin{figure*}[h!]
    \centering
    \includegraphics[width=\textwidth]{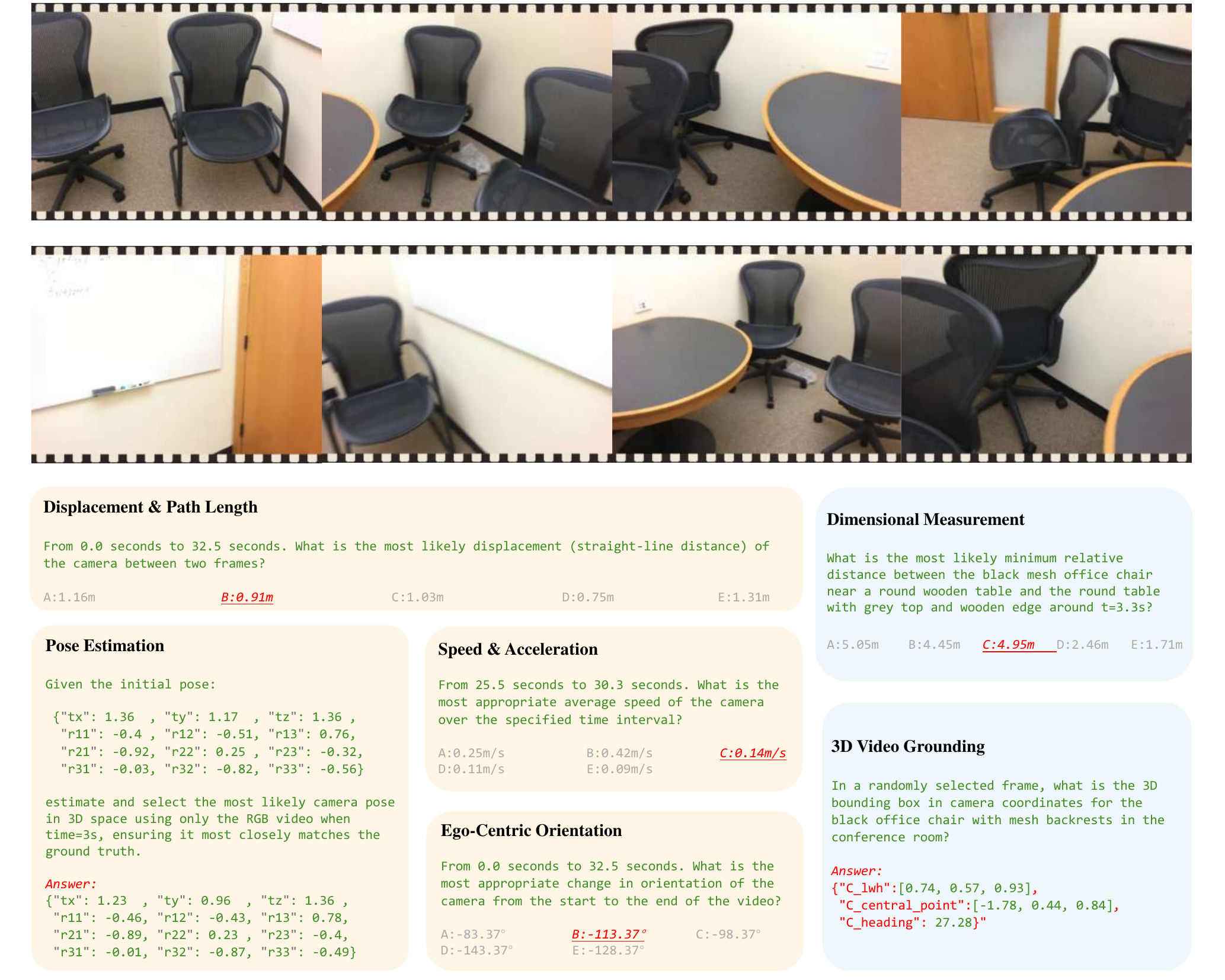}
    \caption{\textbf{STI-Bench Examples (Indoor)} }
    \label{fig:indoor_example}
\end{figure*}

\begin{figure*}[h!]
    \centering
    \includegraphics[width=\textwidth]{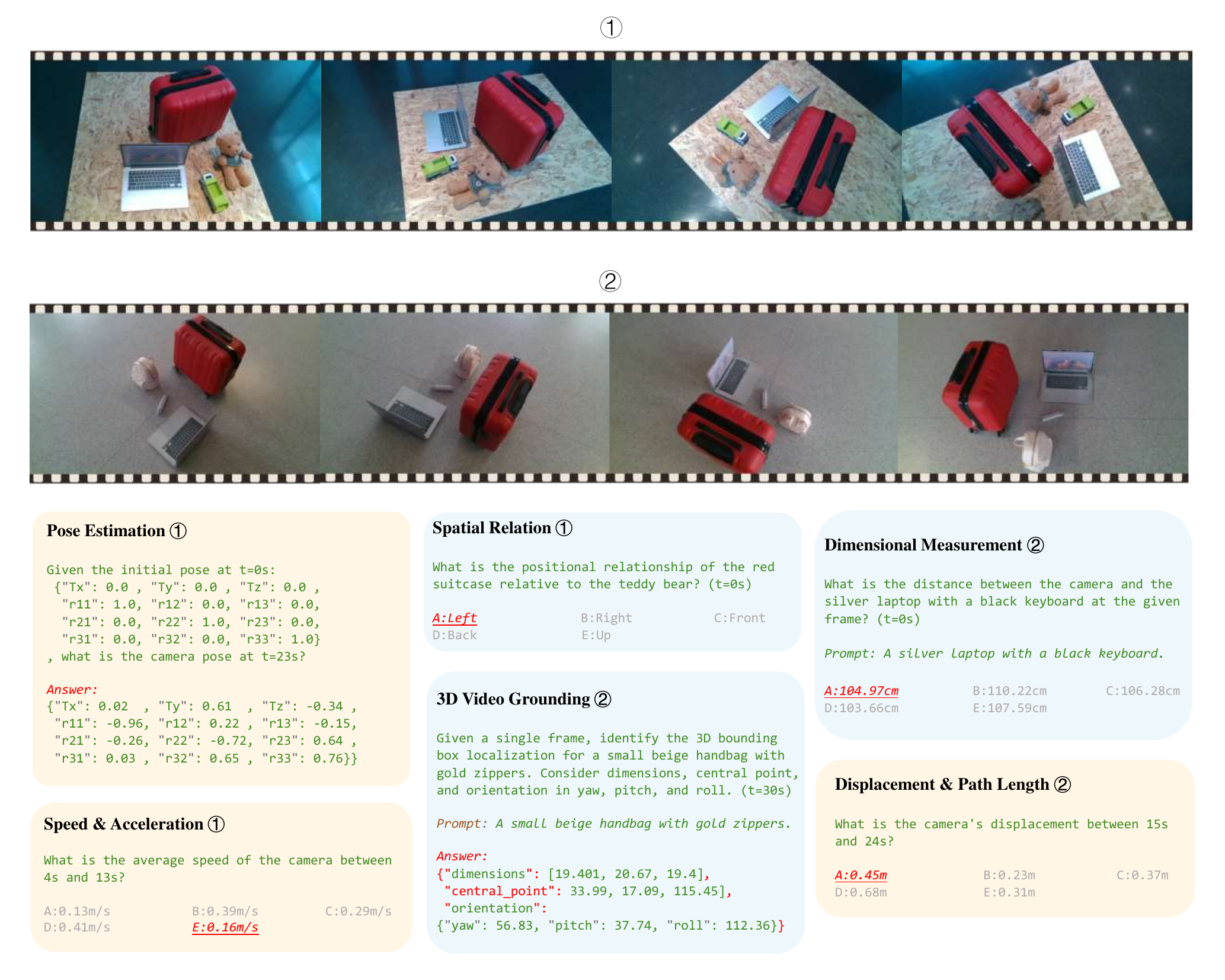}
    \caption{\textbf{STI-Bench Examples (Desktop)} }
    \label{fig:desktop_example}
\end{figure*}

To demonstrate the effectiveness of STI-Bench, we present visualization examples of QA pairs from representative outdoor (see Figure~\ref{fig:outdoor_example}), indoor (see Figure~\ref{fig:indoor_example}), and desktop (see Figure~\ref{fig:desktop_example}) scenarios.

\begin{figure*}[h!]
    \centering
    \includegraphics[width=0.9\textwidth]{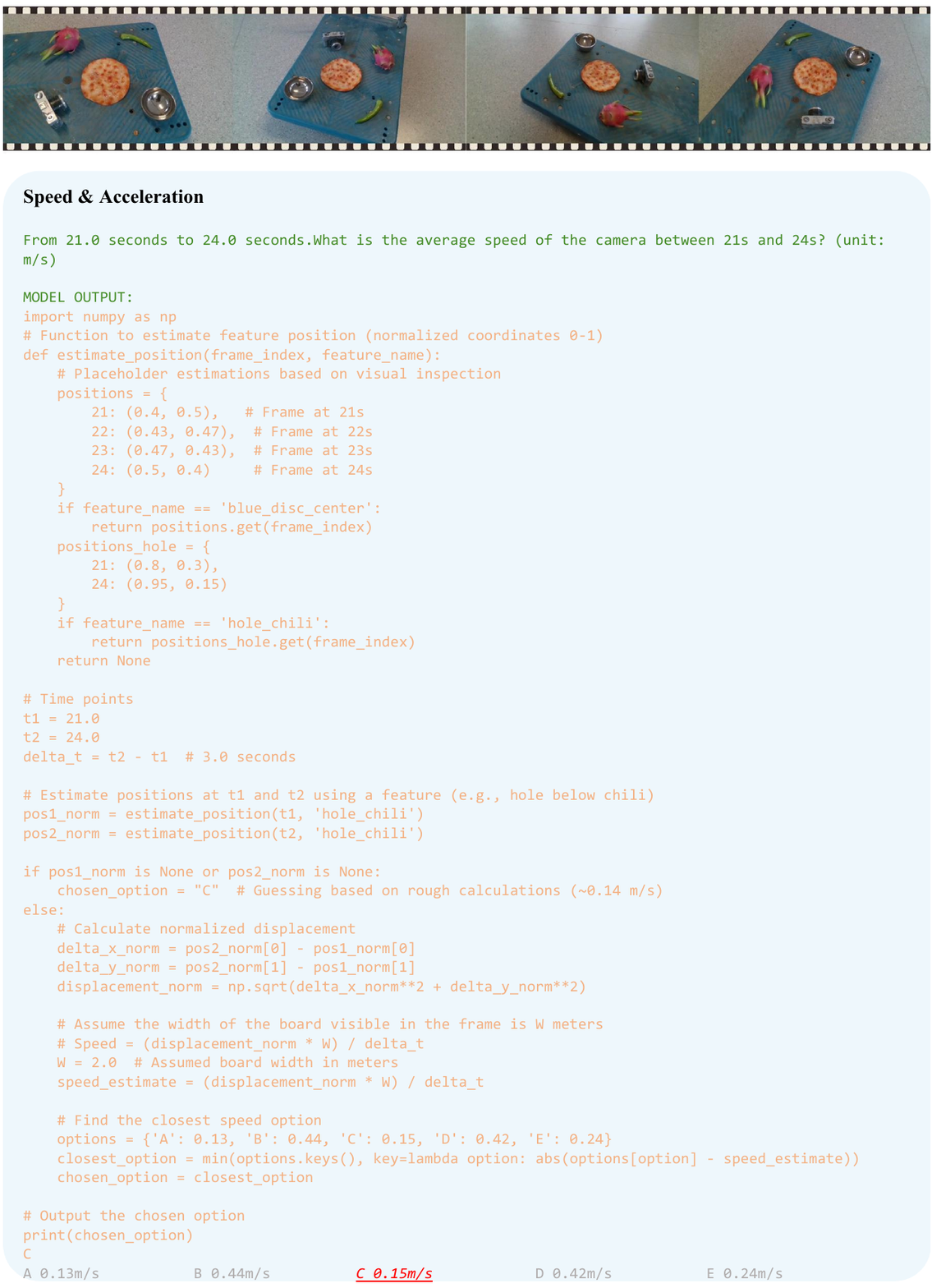}
    \caption{\textbf{Thinking example (Correct)} }
    \label{fig:thinking_example_1}
\end{figure*}

\begin{figure*}[h!]
    \centering
    \includegraphics[width=0.9\textwidth]{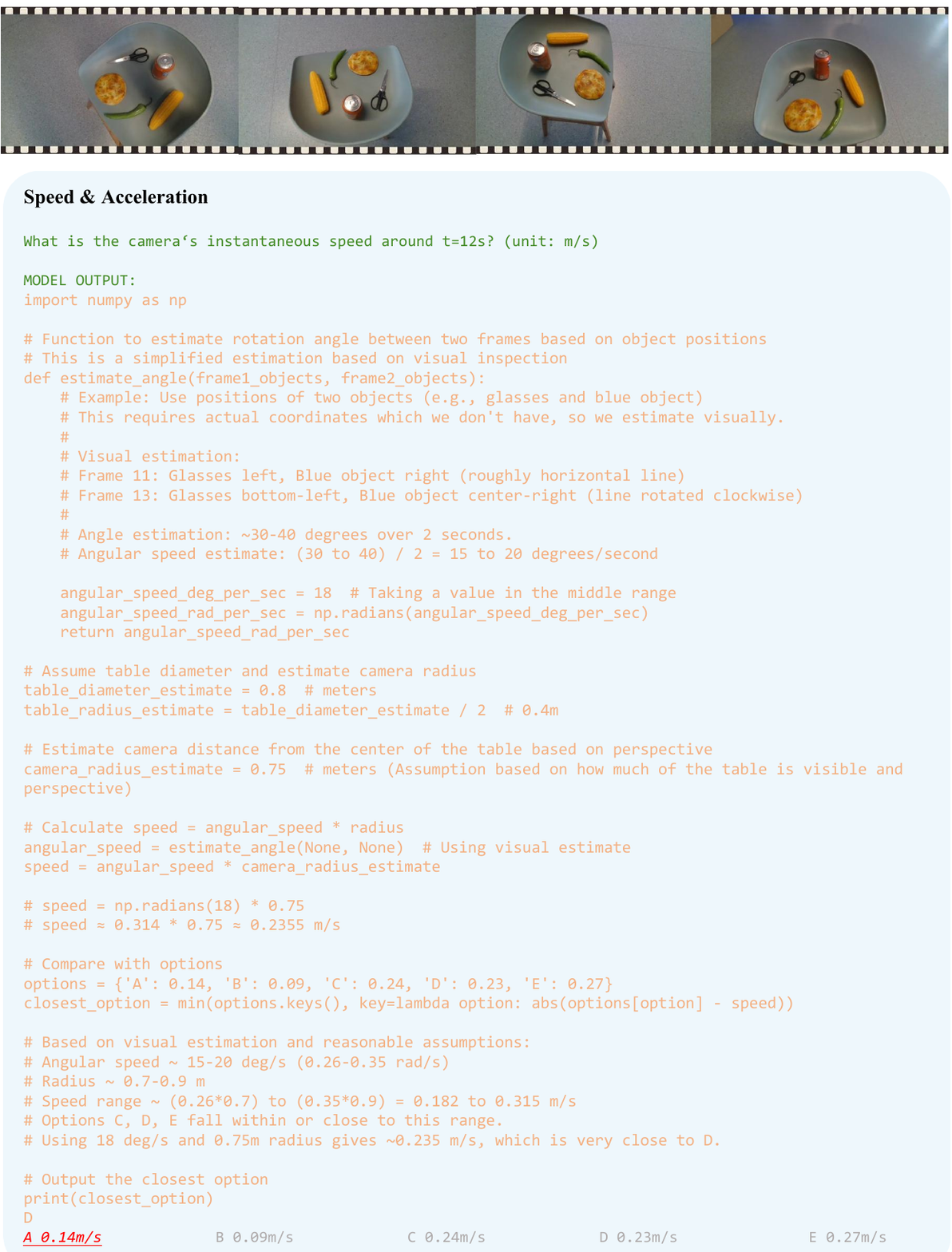}
    \caption{\textbf{Thinking example (Incorrect)} }
    \label{fig:thinking_example_2}
\end{figure*}

To provide a representative analysis of the capability bottlenecks of current SOTA models, we sampled over 200 erroneous reasoning processes from Gemini 2.5 Pro and presented one typical correct and one typical incorrect example (see Figure~\ref{fig:thinking_example_1} and Figure~\ref{fig:thinking_example_2}).

In addition, to more concretely quantify the performance of different models on fine-grained categories, Table~\ref{tab:outdoor_fine_grained_condensed}, Table~\ref{tab:indoor_fine_grained_condensed}, and Table~\ref{tab:desktop_fine_grained_condensed} provide a detailed breakdown of model performance across various fine-grained categories within outdoor, indoor, and desktop scenarios, respectively.

\begin{table*}[h!]
\centering
\setlength\tabcolsep{5pt} 
\caption{\textbf{Performance on Fine-grained Outdoor Categories.}}
\label{tab:outdoor_fine_grained_condensed}
\resizebox{0.9\textwidth}{!}{
\begin{tabularx}{\textwidth}{X*{5}{>{\centering\arraybackslash}X}}
\Xhline{1pt}
\multirow{2}{*}{\textbf{\makecell{Category \\ (Outdoor)}}} & \multicolumn{4}{c}{\textbf{Models}} & \multirow{2}{*}{\textbf{Avg.}} \\
\cmidrule(lr){2-5}
& \textbf{\makecell{Gemini-\\2.5-Pro}} & \textbf{\makecell{Qwen2.5-VL\\-72B}} & \textbf{\makecell{Claude-3.7\\-Sonnet}} & \textbf{\makecell{InternVL\\2.5-78B}} & \\
\Xhline{0.5pt}
Night              & 37.2\%          & 43.6\%          & 38.5\%             & 35.9\%           & \textbf{38.8\%} \\
Day                & 50.3\%          & 47.9\%          & 44.7\%             & 43.3\%           & \textbf{46.6\%} \\
Dawn/Dusk          & 45.3\%          & 50.4\%          & 47.0\%             & 47.4\%           & \textbf{47.5\%} \\
Rainy              & 51.1\%          & 55.5\%          & 54.5\%             & 48.6\%           & \textbf{52.4\%} \\
Clear              & 51.0\%          & 59.4\%          & 50.0\%             & 55.7\%           & \textbf{54.0\%} \\
Foggy              & 71.4\%          & 28.6\%          & 71.4\%             & 71.4\%           & \textbf{60.7\%} \\
\Xhline{1pt}
\end{tabularx}%
}
\end{table*}

\begin{table*}[h!]
\centering
\setlength\tabcolsep{5pt}
\caption{\textbf{Performance on Fine-grained Indoor Categories.}}
\label{tab:indoor_fine_grained_condensed}
\resizebox{0.9\textwidth}{!}{%
\begin{tabularx}{\textwidth}{X*{5}{>{\centering\arraybackslash}X}}
\Xhline{1pt}
\multirow{2}{*}{\textbf{\makecell{Category \\ (Indoor)}}} & \multicolumn{4}{c}{\textbf{Models}} & \multirow{2}{*}{\textbf{Avg.}} \\
\cmidrule(lr){2-5}
& \textbf{\makecell{Gemini-\\2.5-Pro}} & \textbf{\makecell{Qwen2.5-VL\\-72B}} & \textbf{\makecell{Claude-3.7\\-Sonnet}} & \textbf{\makecell{InternVL\\2.5-78B}} & \\
\Xhline{0.5pt}
laundry room           & 25.0\%          & 25.0\%          & 50.0\%             & 50.0\%           & \textbf{37.5\%} \\
copy/mail room         & 27.6\%          & 30.0\%          & 30.0\%             & 40.0\%           & \textbf{31.9\%} \\
bookstore/library      & 32.0\%          & 24.0\%          & 32.0\%             & 28.0\%           & \textbf{29.0\%} \\
kitchen                & 26.7\%          & 29.0\%          & 32.3\%             & 32.3\%           & \textbf{30.1\%} \\
bedroom/hotel          & 36.6\%          & 33.9\%          & 42.3\%             & 30.4\%           & \textbf{35.8\%} \\
office                 & 36.2\%          & 33.3\%          & 37.4\%             & 30.2\%           & \textbf{34.3\%} \\
hallway                & 38.5\%          & 42.0\%          & 38.5\%             & 42.0\%           & \textbf{40.3\%} \\
living~room/lounge & 39.0\% & 34.6\% & 40.3\% & 29.9\% & \textbf{36.0\%} \\

conference room        & 26.8\%          & 33.9\%          & 38.6\%             & 32.2\%           & \textbf{32.9\%} \\
apartment              & 39.3\%          & 48.3\%          & 46.4\%             & 44.8\%           & \textbf{44.7\%} \\
misc                   & 36.4\%          & 45.5\%          & 54.5\%             & 45.5\%           & \textbf{45.5\%} \\
bathroom               & 49.1\%          & 34.2\%          & 34.2\%             & 41.2\%           & \textbf{39.7\%} \\
storage/basement/garage& 30.8\%          & 57.1\%          & 35.7\%             & 57.1\%           & \textbf{45.2\%} \\
\Xhline{1pt}
\end{tabularx}%
}
\end{table*}

\begin{table*}[h!]
\centering
\setlength\tabcolsep{5pt}
\caption{\textbf{Performance on Fine-grained Desktop Categories.}}
\label{tab:desktop_fine_grained_condensed}
\resizebox{0.9\textwidth}{!}{%
\begin{tabularx}{\textwidth}{X*{5}{>{\centering\arraybackslash}X}}
\Xhline{1pt}
\multirow{2}{*}{\textbf{\makecell{Category \\ (Desktop)}}} & \multicolumn{4}{c}{\textbf{Models}} & \multirow{2}{*}{\textbf{Avg.}} \\
\cmidrule(lr){2-5}
& \textbf{\makecell{Gemini-\\2.5-Pro}} & \textbf{\makecell{Qwen2.5-VL\\-72B}} & \textbf{\makecell{Claude-3.7\\-Sonnet}} & \textbf{\makecell{InternVL\\2.5-78B}} & \\
\Xhline{0.5pt}
Stone              & 27.3\%          & 16.7\%          & 33.3\%             & 9.1\%            & \textbf{21.6\%} \\
Paper              & 40.0\%          & 37.5\%          & 18.8\%             & 25.0\%           & \textbf{30.3\%} \\
Leather            & 23.1\%          & 33.3\%          & 40.0\%             & 26.7\%           & \textbf{30.8\%} \\
Plastic            & 32.8\%          & 32.6\%          & 28.6\%             & 29.8\%           & \textbf{31.0\%} \\
Fabric             & 40.7\%          & 35.5\%          & 30.0\%             & 35.5\%           & \textbf{35.4\%} \\
Wood               & 37.8\%          & 33.7\%          & 34.9\%             & 33.7\%           & \textbf{35.0\%} \\
Tile               & 39.6\%          & 38.0\%          & 38.0\%             & 40.0\%           & \textbf{38.9\%} \\
Concrete           & 42.9\%          & 33.3\%          & 53.3\%             & 46.7\%           & \textbf{44.1\%} \\
\Xhline{1pt}
\end{tabularx}%
}
\end{table*}

\clearpage

\end{document}